\documentclass{article}

\usepackage{graphicx, amsmath, fullpage, amssymb, amsthm, amsfonts, color, algorithm}
\usepackage{enumerate}

\newtheorem{theorem}{Theorem}[section]
\newtheorem{lemma}[theorem]{Lemma}

\theoremstyle{definition}
\newtheorem{definition}[theorem]{Definition}

\theoremstyle{remark}

\numberwithin{equation}{section}
\usepackage[colorlinks,
            linkcolor=red,
            anchorcolor=red,
            citecolor=blue
            ]{hyperref}

\newcommand{\R}{\mathbb R}
\newcommand{\N}{\mathbb N}

\renewcommand{\epsilon}{\varepsilon}

\DeclareMathOperator*{\argmax}{arg\,max}

\begin{document}

\title{Spectral gap-based deterministic tensor completion}

% \author[1]{Kameron Decker Harris\thanks{\url{harri267@wwu.edu}}}
% \author[2]{Oscar L\'opez\thanks{\url{lopezo@fau.edu}}}
% \author[1]{Angus Read\thanks{\url{reada2@wwu.edu}}}
% \author[3]{Yizhe Zhu\thanks{\url{yizhe.zhu@uci.edu}}}
% \affil[1]{Department of Computer Science, Western Washington University, Bellingham, WA, USA}
% \affil[2]{Harbor Branch Oceanographic Institute, Florida Atlantic University, Fort Pierce, FL, USA}
% \affil[3]{Department of Mathematics, University of California, Irvine, CA, USA}
% %\affil[ ]{\small \url{harri267@wwu.edu}, \url{lopezo@fau.edu}, \url{reada2@wwu.edu}, \url{yizhe.zhu@uci.edu}}

\author{Kameron Decker Harris \\
  Department of Computer Science\\
  Western Washington University\\
  Bellingham, WA, USA\\
  \url{harri267@wwu.edu}
  \and
  Oscar L\'opez\\
  Harbor Branch Oceanographic Institute\\
  Florida Atlantic University\\
  Fort Pierce, FL, USA\\
  \url{lopezo@fau.edu}
  \and
  Angus Read\\
  Department of Computer Science\\
  Western Washington University\\
  Bellingham, WA, USA\\
  \url{reada2@wwu.edu}
  \and
  Yizhe Zhu\\
  Department of Mathematics\\
  University of California, Irvine\\
  Irvine, CA, USA\\
\url{yizhe.zhu@uci.edu}}
  
% \author{
% \IEEEauthorblockN{Kameron Decker Harris}
% \IEEEauthorblockA{\textit{Dept.\ of Computer Science} \\
% \textit{Western Washington University}\\
% Bellingham, WA, USA \\
% harri267@wwu.edu}
% \and
% \IEEEauthorblockN{Oscar L\'opez}
% \IEEEauthorblockA{\textit{Harbor Branch Oceanographic Institute} \\
% \textit{Florida Atlantic University}\\
% Fort Pierce, FL, USA \\
% lopezo@fau.edu}
% \and
% \IEEEauthorblockN{Angus Read}
% \IEEEauthorblockA{\textit{Dept.\ of Computer Science} \\
% \textit{Western Washington University}\\
% Bellingham, WA, USA \\
% reada2@wwu.edu}
% %\and
% \linebreakand % centers Yizhe's block

% \IEEEauthorblockN{Yizhe Zhu}
% \IEEEauthorblockA{\textit{Dept.\ of Mathematics} \\
% \textit{University of California, Irvine}\\
% Irvine, CA, USA \\
% yizhe.zhu@uci.edu}
% }
\date{\today}

\maketitle

\begin{abstract}

Tensor completion is a core machine learning algorithm used in recommender systems and other domains with missing data.
While the matrix case is well-understood, theoretical results for tensor problems are limited, particularly when the sampling patterns are deterministic.
Here we bound the generalization error of the solutions of two tensor completion methods, Poisson loss and atomic norm minimization, providing tighter bounds in terms of the target tensor rank.
If the ground-truth tensor is order $t$ with CP-rank $r$, the dependence on $r$ is improved from $r^{2(t-1)(t^2-t-1)}$ in \cite{harris2021deterministic} to $r^{2(t-1)(3t-5)}$. The error in our bounds is deterministically controlled by the spectral gap of the sampling sparsity pattern. We also prove several new properties for the atomic tensor norm, reducing the rank dependence from $r^{3t-3}$ in \cite{ghadermarzy2018} to $r^{3t-5}$ under random sampling schemes.
A limitation is that atomic norm minimization, while theoretically interesting, leads to inefficient algorithms.
However, numerical experiments illustrate the dependence of the reconstruction error on the spectral gap for the practical max-quasinorm, ridge penalty, and Poisson loss minimization algorithms.
This view through the spectral gap is a promising window for further study of tensor algorithms.
\end{abstract}
\section{Introduction}

In many situations, incomplete data measurements are the rule due to data corruption, impracticality, or impossibility of filling in all information that is desired.
Tensor completion is a general technique used to fill in missing multivariate data using assumptions of prior structure.
It is the natural generalization of matrix completion, made famous as a winning solution to the Netflix prize \cite{bennett2007netflix}: 
A system to recommend films to users was built by filling in a low-rank matrix of scores for each pair of movies by users.

In tensor completion, we seek to fully or partially recover the  entries of an unknown order $t$ tensor (i.e., a multidimensional $n \times \cdots \times n$ array with $t \geq 2$ indices) $T$ given some limited set of observations
$\{ T_e \}_{e \in E}$.
The ground truth is assumed to have CP-rank $r$ \cite{kolda2009tensor}. Mathematical analysis of such methods often assumes that the observed entries in the tensor $E \subseteq [n]^t$ are random, and the best known polynomial-time algorithms require $\tilde O(n^{t/2})$ sample complexity for an order $t$ tensor \cite{jain2014provable,montanari2018spectral,xia2019polynomial,cai2019nonconvex}.
On the other hand, a number of papers have studied deterministic sampling for matrix completion \cite{heiman2014deterministic,brito2018spectral,bhojanapalli2014universal,burnwal2020deterministic,chatterjee2020deterministic,foucart2020weighted}.
For tensors, only a few recent results have studied error bounds in the deterministic setting.
\cite[Theorem 3.1]{chao2021hosvd} gave a general bound for non-uniform Frobenius norm error for any deterministic sampling pattern with an NP-hard algorithm. However, the dependence on rank in the bound is not specified. 
An HOSVD method was also analyzed in \cite{chao2021hosvd} for low Tucker rank tensor completion.

Closest to the work we present here,
\cite{harris2021deterministic} found bounds for minimizing a quasinorm penalized problem \cite{ghadermarzy2018} in terms a {\em spectral gap}.
For regular graphs where each vertex has the same degree,  {\em smaller second eigenvalue corresponds to larger spectral gap}. 
Viewing the sampling pattern as the adjacency tensor of a $t$-uniform  hypergraph $H$  constructed from a regular graph $G$, they showed that the second eigenvalue  of $G$ could be used to control the reconstruction error.
Expander graphs, which have the smallest possible second eigenvalue/largest spectral gap, are thus optimal for that theory. In addition, using the definition of the second eigenvalue for the adjacency tensor of a hypergraph \cite{friedman1995second}, they showed that the for any $t$-uniform hypergraph (not necessarily regular) as a sampling pattern, its second eigenvalue  could also  be used to control the reconstruction error, but with a larger sample complexity bound.

Our main contribution is a collection of theoretical results suggesting that the spectral gap influences the quality of reconstruction.
We first show how the rank dependence of previous generalization bounds \cite{harris2021deterministic} can be tightened if the optimization is done with a different atomic norm penalty.
Several properties of this atomic norm are proven, improving upon the results in \cite{ghadermarzy2018} where the atomic norm was introduced.
We also extend the analysis of tensor count data with a Poisson likelihood model \cite{lopez2022zero} to deterministic observations.
Finally, we present numerical experiments to showcase how a larger spectral gap leads to less error.
These results together provide strong evidence that the spectral gap of the observation mask is important for controlling the error of tensor completion.
Our notation and some basic definitions are given in App.~\ref{sec:notation}.

\section{Theoretical results}
\subsection{Tensor atomic norm properties}
In order to connect the expansion properties of  the sampling pattern hypergraph 
to the error of the algorithms,
we will work with sign tensors. 
A sign tensor $S$ has all entries equal to $+1$ or $-1$, i.e.\
$S \in \bigotimes_{i=1}^t \{ \pm 1\}^{n_i}$.
The {\bf sign rank} of a sign tensor $S$ is defined as
\begin{align}
    \mathrm{rank}_\pm (S)
    = \inf
    \left\{ r \; \Big| \; 
    S = \sum_{i=1}^r s^{(1)}_i \circ \cdots \circ s^{(t)}_i\right\},
    \label{eq:sign_rank}
\end{align}
 where  $s_i^{(j)} \in \{ \pm 1 \}^{n_i}, i \in [r], j \in [t]$. We define the {\bf atomic norm} for a tensor $T$ as
\begin{align}\label{eq:def_atomic}
    \| T \|_\pm & = 
    \inf \left\{ \sum_{i=1}^r |\alpha_i|
    \; \Big| \; T = \sum_{i=1}^r \alpha_i S_i 
    \right\}, 
\end{align}
 where 
    $\alpha_i \in \R, \,
      S_i \in  \bigotimes_{i=1}^t\{\pm 1\}^{n_i}, i\in [r] $.
This is called \textit{tensor atomic-$M$ norm} in \cite{ghadermarzy2018}. 
In \eqref{eq:def_atomic}, the ``atoms'' are  simple sign tensors.

Note that the set of all rank-1 sign tensors forms a basis for
$\bigotimes_{i=1}^t \R^{n}$, 
so this decomposition into rank-1 sign tensors is always possible;
furthermore, this is a norm for tensors and matrices
\cite{ghadermarzy2018,heiman2014deterministic}, and it is commonly used in compressed sensing and matrix completion \cite{chandrasekaran2012convex,chi2020harnessing}.
The following results give several useful properties of the tensor atomic norm, which will be used in our tensor completion analysis
(proofs in Apps.~\ref{pf:atomic-properties} and \ref{pf:rank-1}, respectively):
\begin{theorem}
\label{thm:atomic-properties}
Let $T \in \bigotimes_{i=1}^t \R^{n_i}$
and $S \in \bigotimes_{i=1}^t \R^{m_i}$.
The following atomic norm properties hold:
\begin{enumerate}
    \item $ \| T_{I_1, \ldots, I_t} \|_{\pm}
            \leq
            \| T \|_{\pm}$\label{claim:1}
    for any subsets
    $I_i \subseteq [n_i]$.
    \item $\| T \otimes S \|_{\pm}
            \leq
            \| T \|_{\pm} \| S \|_{\pm}$. \label{claim:2}
    \item $\| T *S \|_{\pm} \leq \| T \otimes S \|_{\pm}$,
    where $T,S\in  \bigotimes_{i=1}^t \R^{n_i}$.\label{claim:3}
\item $
\| T *T  \|_{\pm} \leq \| T \|_{\pm}^2.$\label{claim:4}
\end{enumerate}
\end{theorem}

\begin{lemma}\label{lem:rank-1}
   Let $u_i\in \mathbb R^{n_i}$ with 
   $\|u_i\|_{\infty}\leq 1$ for $i\in [t]$ and rank-1
   $T=u_1\circ u_2\circ \cdots \circ  u_t$. Then
   $\|T\|_{\pm}\leq 1.$
\end{lemma}

We also need to compare the atomic norm of a tensor $T$ and its CP-rank to obtain rank-dependent bounds.
The following Theorem improves the rank dependence in \cite[Theorem 7]{ghadermarzy2018} by a factor of $r$ (proof in App.~\ref{pf:atomic_rank}):
\begin{theorem}\label{thm:atomic_rank}
   Let $T \in \bigotimes_{i=1}^t \R^{n_i}$  and $\mathrm{rank}(T)=r$. Then 
   \begin{align}\label{eq:improve_atomic}
   |T|_{\infty} \leq  \|T\|_{\pm} \leq K_G\sqrt{r^{3t-5}} |T|_{\infty},
   \end{align}
   where $K_G\leq 1.783$ is  Grothendieck's constant over $\mathbb R$.
\end{theorem}

In contrast, for the max-quasinorm (see App.~\ref{sec:notation}), it was shown in \cite{harris2021deterministic} that 
\begin{equation*}
|T|_\infty
\leq \| T \|_\mathrm{max} 
\leq 
\sqrt{r^{t^2-t-1}} 
~
|T|_\infty.
\end{equation*}
Since $3t-5\leq t^2-t-1$ for all integers $t\geq 2$, atomic norm-based analysis and optimization will yield a better rank dependence in the generalization error bound.

While the main focus of this paper is on deterministic sampling patterns, we note that Theorem \ref{thm:atomic_rank} can be used to improve upon results in the literature that consider random sampling schemes. In \cite{ghadermarzy2018}, the authors introduce the atomic norm to show that $O\left(nr^{3t-3}\right)$ entries chosen randomly are sufficient to provide an accurate approximation of a rank $r$ tensor in $\bigotimes_{i=1}^t \R^{n}$. Using our main result, Theorem 8 in \cite{ghadermarzy2018} can be applied with $R = K_G\sqrt{r^{3t-5}} |T|_{\infty}$ to reduce the sampling complexity to $O\left(nr^{3t-5}\right)$. To the best of our knowledge, the results here provide the best sampling complexity to date for tensors of general order under random and deterministic sampling patterns.

\subsection{Deterministic tensor completion}

Equipped with the properties of the atomic norm proved in Theorem \ref{thm:atomic-properties}, we give an improved generalization error bound for deterministic tensor completion
(proof in App.~\ref{pf:main3}):
\begin{theorem}
\label{thm:main3}
Given a hypercubic tensor $T$ of order $t$, 
reveal its entries according to a  $t$-uniform  $t$-partite hypergraph
$H=(V,E)$ with $V=V_1\cup \cdots \cup V_t$, $|V_1|=\cdots=|V_t|=n$, and second eigenvalue $\lambda_2(H)$.
Let $\hat{T}$ satisfy
\begin{align*}
\hat{T} &= \arg \min_{T'} \| T' \|_{\pm}\\
&\mbox{\quad such that \quad
$T'_{e} = T_{e}$
\quad for all\quad  $e \in E$.}
\end{align*}
Then the following 
bound holds:
\begin{align} \label{eq:errorbound15}
\frac{1}{n^t}
\|\hat{T}-T\|_F^2
&\leq
\frac{2^{t+2}n^{t/2}\lambda_2(H)}{|E|}\|T\|_{\pm}^2, 
\end{align}
where 
$
%\begin{align}
%\label{eq:lambda2H}
    \lambda_{2}(H)=\left\|T_H-\frac{|E|}{n^t} J\right\|,
%\end{align}
$
$J$ is the all-ones tensor, and $T_H$ is the adjacency tensor of $H$ such that 
$T_{i_1,\dots, i_t}=\mathbf{1}\{ (i_1,\dots, i_t)\in E\}$.
\end{theorem}
The above result is for any $t$-uniform, $t$-partite hypergraphs, and the error bound depends on the second eigenvalue of their adjacency tensors. 
Concentration for the  second eigenvalue of adjacency tensors for random hypergraphs was considered in \cite{friedman1995second,zhou2019sparse}.

\begin{figure}
\centering
\includegraphics[width=0.5\linewidth]{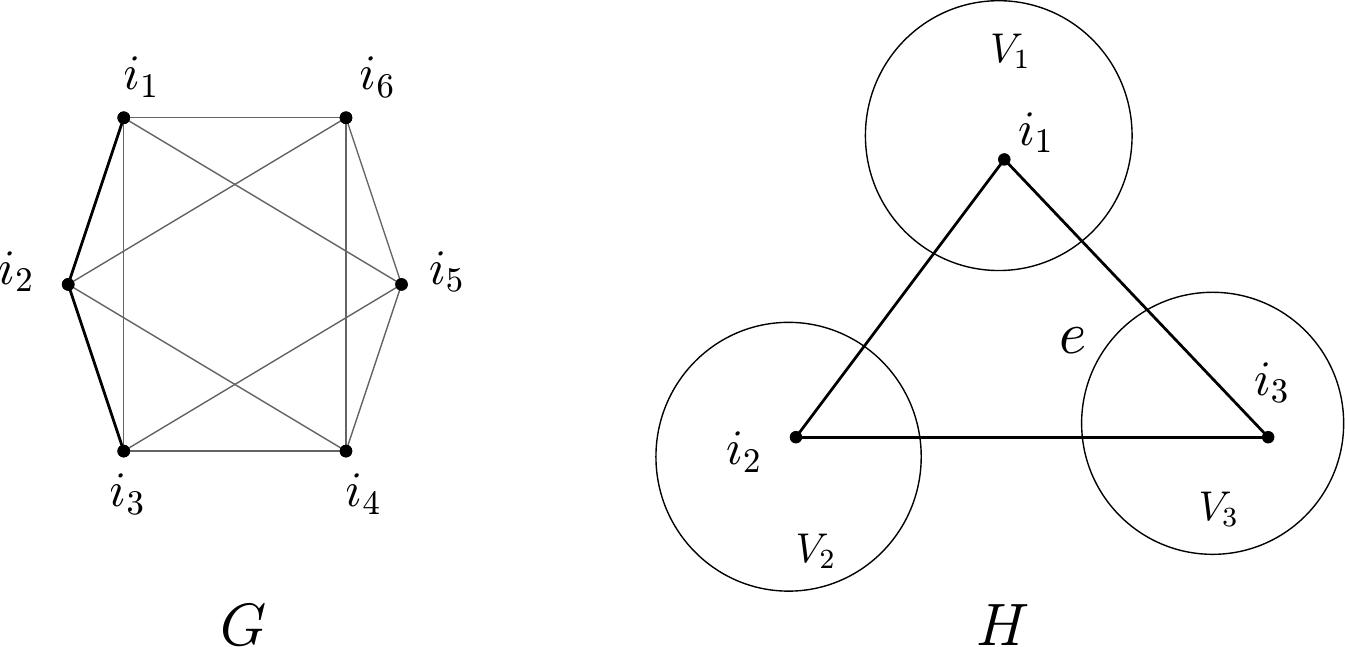}
\caption{Lifting a graph $G$ into a hypergraph $H$ when $t=3$:
We depict a 4-connected ring base graph $G$ on the left and a single edge in the hypergraph $H$ on the right.
$(i_1, i_2, i_3)$
forms an hyperedge $e$ in $H$ if and only if 
$(i_1,i_2,i_3)$ is a walk in $G$.
% For convenience, we denote that hyperedge 
% with the tuple
% $(i_1,i_2,i_3)$.
}
\label{fig:graph}
\end{figure}

Computing $\lambda_2(H)$ is costly since the tensor spectral norm is NP-hard \cite{hillar2013most}.
However, we can ``lift'' a graph into a hypergraph in a way that gives a bound in terms of graph eigenvalues.
Let $G=(V(G), E(G))$ be a connected $d$-regular graph on $n$ vertices
with the second largest  eigenvalue (in absolute value)
$\lambda\in (0,d)$. 
The following construction of  a $t$-partite, $t$-uniform, $d^{t-1}$-regular hypergraph 
$H=(V,E)$ from $G$ was given in \cite{harris2021deterministic}.
See Fig~\ref{fig:graph} for an illustration of the ``edge lifting'' operation from a regular graph to a regular hypergraph graph when $t=3$ and $d=4$.

\begin{definition}[Edge lifting for regular hypergraphs]\label{eq:def_regular}
Let $V=V_1\cup V_2\cup \cdots \cup V_t$ 
be the disjoint union of $t$ vertex sets
such that $|V_1| = \cdots = |V_t|=n$. 
% To be precise,
% we label all the vertices by 
% $v_{i}^j\in V_i$ for all $1\leq i\leq t$ and $1\leq j\leq n$.
The hyperedges of $H$ correspond to all walks  of length $t-1$ in $G$:
$(v_1 ,\dots, v_t)$ 
is a hyperedge in $H$ if and only if 
$(i_1,\dots, i_{t})$ is a walk of length $t-1$ in $G$.
\end{definition}

\begin{theorem}\label{thm:main_reg}
Given a hypercubic tensor $T$ of order $t$, 
reveal its entries according to a 
$t$-partite, $t$-uniform,
$d^{t-1}$-regular hypergraph
$H=(V,E)$
lifted from a $d$-regular graph $G$ of size $n$
with second eigenvalue (in absolute value) $\lambda\in (0,d)$.
Then solving 
\begin{align}\label{eq:opt}
\hat{T} &= \arg \min_{T'} \| T' \|_{\pm} \notag\\
&\mbox{\quad such that \quad
$T'_{e} = T_{e}$
\quad for all\quad  $e \in E$}
\end{align}
will result in the following mean squared error bound:
\begin{align}\label{eq:errorbound}
\frac{1}{n^t}
\|\hat{T}-T\|_F^2
&\leq 2^t(2t-3)\frac{\lambda}{d} \|T\|_{\pm}^2 \notag \\
&\leq 2^t(2t-3)\frac{\lambda}{d}K_G^2r^{3t-5}|T|_{\infty}^2.
\end{align}
\end{theorem}

Theorem~\ref{thm:main_reg} is proven in App.~\ref{pf:main_reg}.
Suppose we lift an expander graph $G$ with $\lambda=O(\sqrt{d})$ and $t,|T|_{\infty}=O(1)$. 
In order to have the right hand side of \eqref{eq:errorbound} bounded by $\epsilon$, we need to take
\begin{equation}\label{sampcomp}
    |E|=O\left(\frac{nr^{2(t-1)(3t-5)}}{\epsilon^{2(t-1)}}\right)
\end{equation}
many samples.
The sample complexity in \cite{harris2021deterministic} is 
$O\left(\frac{nr^{2(t-1)(t^2-t-1)}}{\epsilon^{2(t-1)}}\right)$.
Optimizing the atomic norm instead of the max-quasinorm yields a better bound, although it requires costly integer programming. 

\section{Poisson Tensor Regression}

We may also use the atomic norm properties to obtain error bounds in the case of noisy tensor completion. 
In this scenario, we seek to estimate a parametric tensor $T$ from noisy, incomplete observations so that $T$ is never observed directly.
We specifically consider the Poisson tensor completion problem \cite{lopez2022zero}, where we observe count data 
$X \in \bigotimes_{i=1}^t \N_+^{n_i}$ 
obeying
\begin{equation} \label{Pdist}
    X_e \sim \mbox{Poisson}(T_e) \ \ \mbox{for all} \ \ e\in E,
\end{equation}
where $T \in \bigotimes_{i=1}^t [\beta,\alpha]^{n_i}$ specifies the range of possible Poisson parameters (with $0<\beta\leq\alpha$). 
Given $X_e$ for $e\in E$, we approximate $T$ via a Poisson maximum likelihood estimator
\begin{equation} \label{Pest}
    \hat{T} = \argmax_{Z\in S_r(\beta,\alpha)}
    \sum_{e\in E} X_e \log(Z_e) - Z_e
\end{equation}
where our parametric tensor search space is
\[
S_r(\beta,\alpha) = \Bigg\{Z \in \bigotimes_{i=1}^t [\beta,\alpha]^{n_i}\ | \ \mbox{rank}(Z)\leq r\Bigg\}.
\]
This leads to our main result for the Poisson regression
(proof in App.~\ref{pf:poisson}):
\begin{theorem}\label{thm:poisson}
Let the hypercubical parameter tensor 
$T$ and observations $X$ be generated as above,
% $T \in \bigotimes_{i=1}^t [\beta,\alpha]^{n}$, 
% let $X \in \bigotimes_{i=1}^t \N_+^{n}$ be generated as in \eqref{Pdist} 
with entries revealed according to a  $t$-uniform,  $t$-partite hypergraph $H=(V,E)$ with $V=V_1\cup \cdots \cup V_t$, $|V_1|=\cdots=|V_t|=n$, and second eigenvalue $\lambda_2(H)$. Then with probability exceeding $1-\frac{2}{|E|}$,
there exists an absolute constant $C>0$ such that $\hat{T}$ satisfies:
\begin{align}\label{eq:Perrorbound2}
\frac{1}{n^t}
\|\hat{T}-T\|_F^2 &\leq\frac{C\alpha^3t^{3/2}\sqrt{nr^{3t-5}}\log_2(n)}{\beta\sqrt{|E|}}\nonumber \\ 
&+ \frac{2^{t+2}n^{t/2}\lambda_2(H)K_G^2\alpha^2r^{3t-5}}{|E|}.
\end{align}

Furthermore, if the entries are revealed according to a 
$t$-partite, $t$-uniform,
$d^{t-1}$-regular hypergraph
$H=(V,E)$
constructed from a $d$-regular graph $G$ of size $n$
with second eigenvalue (in absolute value) $\lambda\in (0,d)$, then with probability exceeding $1-\frac{2}{nd^{t-1}}$,
\begin{align}\label{eq:Perrorbound}
\frac{1}{n^t}
\|\hat{T}-T\|_F^2 &\leq
\frac{C\alpha^3t^{3/2}\sqrt{r^{3t-5}}\log_2(n)}{\beta\sqrt{d^{t-1}}}\notag\\
&+ \alpha^22^t(2t-3)\frac{\lambda}{d}K_G^2r^{3t-5}.
\end{align}

\end{theorem}
If all $r, t, \alpha,\beta$ are independent of $n$, and we take an expander graph with $\lambda=O(\sqrt{d})$, \eqref{eq:Perrorbound} says if $nd^{t-1}$ is $\omega(n\log^2 n)$, then the generalization error goes to $0$ as $n\to\infty$. Fixing $t,\alpha,\beta$, the sample size is 
$O\left(\frac{nr^{2(t-1)(3t-5)}}{\epsilon^{2(t-1)}}+\frac{nr^{3t-5}\log^2(n)}{\epsilon}\right)$ for an $\epsilon$-approximation. 
The result is no longer deterministic due to the random nature of the observed counts $X$, but not the observation mask.
We roughly obtain the same sampling complexity as in Theorem \ref{thm:main_reg} with an extra $O\left(\frac{nr^{3t-5}\log^2(n)}{\epsilon} \right)$ term. 

We end this section by noting that the derived noisy tensor completion error bounds \eqref{eq:Perrorbound2} and \eqref{eq:Perrorbound} are dependent on the Poisson parameter bounds $\beta, \alpha$. Such dependence is typical in the literature under random sampling schemes \cite{PoissonMC,lopez2022zero}. Theoretical results therein exhibit analogous dependence on the distributional parameter space, with numerical experiments that further validate this dependence as $\beta$ and $\alpha$ are varied.
% , where if \eqref{sampcomp} holds, then the right hand side of \eqref{eq:Perrorbound} is bounded by \[\frac{\epsilon^{t-1}\log_2(n)}{\sqrt{r^{(3t-5)(2t-3)}}}+\epsilon.\]

\section{Numerical Experiments}

We now present some numerical experiments that explore our derived error bounds in practice. The goal of this section is to showcase how the accuracy of estimators $\hat{T}$ depends on the spectral gap of the associated hypergraph or ``lifted'' regular graph that specifies the sampled tensor entries. To do so, we will generate subsets of revealed tensor entries $E$ in different manners that vary the second eigenvalue of the associated adjacency tensor or lifted graph. The following experiments keep the cardinality $|E|$ fixed so that only the distribution of the sampled entries contributes to the behavior of the reconstruction errors.

We tested the max-quasinorm minimization algorithm of \cite{harris2021deterministic} with graph lifting for a graph with $d=15$ and random target tensor with $n=100$, $t=3$, and $r=3$.
To vary the eigenvalue $\lambda_2(G)$, we start by creating a $d$-connected ring graph (see Figure~\ref{fig:graph}).
We then perform a number of random edge swaps where we randomly select two edges and switch their endpoints, which preserves the regularity of the graph. This is the classical switch Markov chain for generating random regular graphs \cite{greenhill2022generating}.
As the number of swaps increases, $\lambda_2$ decreases until it stabilizes at approximately $2\sqrt{d-1}$. 
We vary the number of swaps between 0 and 600 to give as wide of a $\lambda_2$ range as possible. 
This graph is then lifted into a hypergraph/tensor mask of observed entries.

\begin{figure}
\centering
\includegraphics[width=0.65\linewidth]{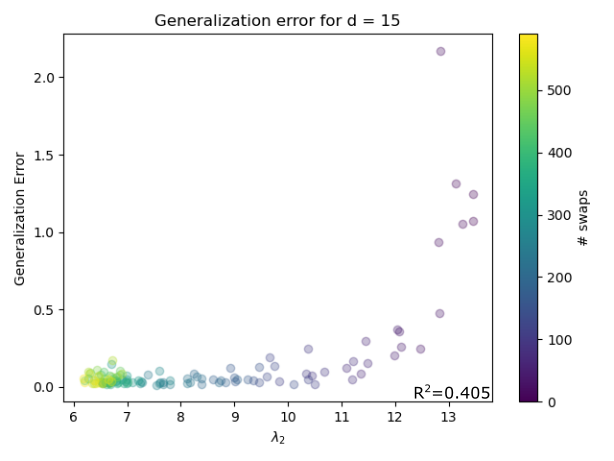}
\includegraphics[width=0.65\linewidth]{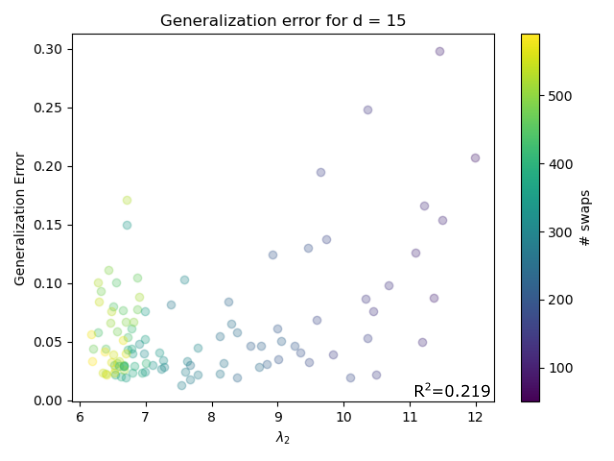} 
\caption{
Numerical experiments show that reconstruction error correlates with $\lambda_2(G)$ using the max-quasinorm minimization method and graph lifting.
This graph has $d=15$ and $n=100$, and the tensor order $t=3$, for 2.25\% of the entries sampled.
Above: The full range of graph eigenvalues by varying edge swaps from 0 to 600.
Below: Linear fits show significant positive correlation even when points with $\lambda_2 > 12$ are removed.
Coefficients of determination $R^2$ for linear fits of error versus $\lambda_2$ are included.
}
\label{fig:fig2}
\end{figure}

Figure~\ref{fig:fig2} shows the results from these experiments.
We observe a significant positive correlation between generalization error and $\lambda_2$, increasing nonlinearly for the larger values of the eigenvalue. 
To check if this pattern persisted when limiting to lower $\lambda_2$ values, we removed the points with large $\lambda_2$ values that were dominating the plot. We can see there still is a positive correlation when the points are removed, providing further evidence that $\lambda_2$ is a determinant of generalization error.
Further details and similar plots for two additional square loss algorithms, as well as regression reports, are included in App.~\ref{sec:num_square} and the supplemental files.

We also experimented with the Poisson regression algorithm \cite{lopez2022zero}.
In this case, we did not use graph lifting but came up with a procedure to generate mask tensors with varying $\lambda_2(H)$ by starting from a grid-like mask and randomly shuffling entries.
The tensor eigenvalue $\lambda_2(H)$ was estimated by a rank-1 fit to $(T_H - \frac{|E|}{n^t}J)$. % as in \eqref{eq:lambda2H}.
We again saw a strong correlation between $\lambda_2$ and error.
Further details and plots of results are shown in App.~\ref{sec:num_poi}.

\section{Conclusion}
We provided an improved analysis of  deterministic tensor completion based on the spectral gap of expander graphs in \cite{harris2021deterministic} and applied the results for Poisson tensor regression. Our new numerical experiments support the dependence of generalization error on
the spectral gap. Our main contribution also improves upon previous results that consider random sampling schemes, providing the best sampling complexity to date for general order tensor completion problems in terms of the CP-rank.

 It would be interesting to see if our analysis can be extended for deterministic non-uniform low CP-rank tensor completion following the line of work \cite{foucart2020weighted,chao2021hosvd}. 
 However, more properties of the tensor atomic or max-quasinorm  are needed. 
 In particular, in the matrix case, we have the following relation between the max-norm and operator norm: $\|A*B\|\leq \|A\|_{\max}\|B\|$ for any two $n\times n$ matrices $A,B$, which is crucial in the proof of \cite[Theorem 15]{foucart2020weighted} and \cite[Theorem B.1]{chao2021hosvd}. 
 Generalizing this inequality for these tensor factorization norms is an interesting question for future work.

Our work also contains some limitations.
First, we did not study the computational complexity of our optimization algorithms, although we expect them to be NP-hard \cite{barak2016noisy}.
Directly minimizing the atomic norm as in Theorem~\ref{thm:main3} requires integer programming and is not efficient in practice.
Finally, while many numerical results show a good correlation with $\lambda_2$, there is significant unexplained variance at a given gap and across algorithms. 
Like many results, these theoretical bounds are not tight enough to quantitatively predict performance, and they are far from the parameter counting lower-bound of $O(nrt)$ for the CP decomposition.

\section*{Acknowledgments}
The authors are listed in alphabetical order.
We would like to thank Daniel Dunlavy for his guidance in setting up the Poisson tensor regression numerical experiments.

O.L. is supported by the Laboratory Directed Research and Development program at Sandia National Laboratories, a multimission laboratory managed and operated by National Technology and Engineering Solutions of Sandia LLC, a wholly owned subsidiary of Honeywell International Inc.\ for the U.S.\ Department of Energy’s National Nuclear Security Administration under contract DE-NA0003525.

Y.Z. is partially supported by  NSF-Simons Research Collaborations on the Mathematical and Scientific
Foundations of Deep Learning.  

\bibliographystyle{plain}
\bibliography{ref}

\onecolumn 

\appendix

\section{Appendix}

\subsection{Notation and definitions}
\label{sec:notation}

We use lowercase symbols $u$ for vectors,
uppercase $U$ for matrices and tensors.
The symbol ``$\circ$'' denotes the outer product of 
vectors, i.e.
$T= u \circ v \circ w$
denotes the order 3, rank-1 tensor with entry 
$T_{i,j,k} = u_i v_j w_k$.
We also use this symbol for the outer product of matrices
as appears in the rank-$r$ decomposition
of a tensor $T = U^{(1)} \circ U^{(2)} \circ U^{(3)}$, 
where each matrix $U^{(i)}$ has $r$ columns,
so that 
$T_{i,j,k} = \sum_{l=1}^r U^{(1)}_{i,l} \, U^{(2)}_{j,l} \, U^{(3)}_{k,l}$,
and
$T = \bigcirc_{i=1}^t U^{(i)}$ 
is shorthand for the same order $t$, rank-$r$ tensor. The symbols $\otimes$ and $*$ denote
Kronecker and Hadamard products, respectively.
We use $\bigotimes_{i=1}^t \R^{n_i}$
for the space of all order $t$ tensors with $n_i$ entries in the $i$-th 
dimension.
We use $1_A \in \R^n$ as the indicator vector of a set $A \subseteq [n]$,
i.e.\ $(1_A)_i = 1$ if $i \in A$ and 0 otherwise. 
For any order $t$ tensor $T\in  \bigotimes_{i=1}^t \R^{n_i}$ and subsets $I_i\subseteq [n_i]$, denote $T_{I_1,\dots,I_t}$ to be the subtensor restricted on the index set $I_1\times \cdots \times  I_t.$
Norms $\| \cdot \|$ are by default the $\ell_2$
norm for vectors and operator norm for matrices and tensors. 
We use the notation $| \cdot |_p$ for
entry-wise $\ell_p$ norms of matrices and tensors. 
% and always include the subscript to avoid confusion
% with set cardinality. \YZ{This might be too long, we can remove some parts}

 Let $T \in \bigotimes_{i=1}^t \R^{n_i}$
    and $S \in \bigotimes_{i=1}^t \R^{m_i}$.
    We define the {\bf Kronecker product}
    of two tensors
    $(T \otimes S) \in \bigotimes_{i=1}^t \R^{n_i m_i}$
    as the tensor with entries
    \[
    (T \otimes S)_{k_1, \ldots, k_t}
    =
    T_{i_1, \ldots, i_t}
    S_{j_1, \ldots, j_t}\]
     for
    $k_1 = j_1 + m_1 (i_1 - 1) , \ldots, 
    k_t = j_t + m_t (i_t - 1)$.
    
    Let $T,S \in \bigotimes_{i=1}^t \R^{n_i}$.
    We define the {\bf Hadamard product}
    of two tensors
    $(T *S) \in \bigotimes_{i=1}^t \R^{n_i}$
    as the tensor with indices
    $
    (T *S)_{i_1, \ldots, i_t}
    =
    T_{i_1, \ldots, i_t}
    S_{i_1, \ldots, i_t} 
    $.

For matrices, 
the most common measure of complexity is the rank.
In the tensor setting, 
there are various definitions of rank 
\cite{kolda2009tensor}.
However, in this paper, 
we will work with  the
\textbf{CP-rank} defined as 
\begin{equation}
    \mathrm{rank}(T) = 
    \min
    \left\{ r \; \Big| \; T = \sum_{i=1}^r u^{(1)}_i \circ \cdots \circ u^{(t)}_i
    \right\} ,
    \label{eq:tensor_rank}
\end{equation}
where each  vectors $u_i^{(j)} \in \R^n$.

In \cite{ghadermarzy2018}, Ghadermarzy, Plan, and Yilmaz
studied tensor completion without reducing it to a matrix case
by minimizing a \textbf{max-quasinorm}
as a proxy for rank.
This is defined as
\begin{equation*}
    \| T \|_\mathrm{max} =
    \min_{T = U^{(1)} \circ \cdots \circ U^{(t)}}
    \prod_{i=1}^t \| U^{(i)} \|_{2,\infty} \; ,
\end{equation*}
where the factorization is a CP decomposition of $T$.

Define the {\bf spectral norm} of a tensor $T$ as
\begin{align}\label{eq:defspecT}
    \|T\|=\sup_{v_1,\dots,v_t\in S^{n-1}} \left|\sum_{i_1,\dots,i_t=1}^n T_{i_1,\dots,i_t}v_1(i_1)\cdots v_t(i_t)\right|,
\end{align}
where $S^{n-1}$ is the unit sphere in $\mathbb R^n$.

A \textbf{$d$-regular graph} on $n$ vertices is a graph where each vertex has the same degree $d$. The \textbf{adjacency matrix} of a graph $G=(V,E)$ is a $|V|\times |V|$ symmetric matrix such that $A_{ij}=\mathbf{1} \{\{i,j\}\in E \}$ for all $i,j\in V$. The \textbf{second eigenvalue} (in absolute value) of $G$, denoted by $\lambda(G)$, is defined as $\lambda=\max\{ |\lambda_2(A)|,|\lambda_n(A)| \}$.

A \textbf{hypergraph} $H=(V,E)$ consists of a set $V$ of vertices and a set $E$ of hyperedges such that each hyperedge is a nonempty set of $V$.  A hypergraph $H$ is \textbf{$k$-uniform} for an integer $k\geq 2$ if every hyperedge $e\in E$ contains exactly $k$ vertices. The \textit{degree} of $i$,  is the number of all hyperedges incident to $i$.
 A hypergraph is \textbf{$d$-regular} if all of its vertices have degree $d$.
 A $k$-uniform hypergraph is \textbf{$k$-partite} if we can decompose the vertex set as a disjoint union $V=V_1\cup\dots \cup V_k$ such that each hyperedge in $E$ contains exactly one vertex in $V_{i}, 1\leq i\leq k$.
The \textbf{adjacency tensor} $T$ of a $k$-uniform hypergraph $H=(V,E)$ on $n$ vertices is a $k$-th order symmetric tensor of size $n$ such that 
$T_{i_1,\dots, i_k}=\mathbf{1} \{ \{i_1,\dots,i_k\} \in E\}$. The second eigenvalue of $H$, denoted by $\lambda_2(H)$, is defined as $
    \lambda_{2}(H)=\left\|T-\frac{|E|}{n^t} J\right\|$, where
$J$ is the all-ones tensor.

\subsection{Proof of Theorem \ref{thm:atomic-properties}}
\label{pf:atomic-properties}
We prove Claims (1--4) in order.
For Claim \eqref{claim:1}, 
write $T$ using the decomposition
which attains the atomic norm,
$T = \bigcirc_{i=1}^t U^{(i)}$
for some $U^{(i)} \in \R^{n_i \times r}, 1\leq i\leq t$.
Then a single entry of $T$ can be written as 
\[T_{i_1, \ldots, i_t} 
= 
\sum_{i=1}^r \alpha_i u^{(1)}_{i_1, i} \cdots u^{(t)}_{i_t, i},\]
where $u_{i_k,i}^{k}\in \{-1,+1\}$ for $1\leq i_k\leq n_k, 1\leq i\leq r, 1\leq k\leq t$,  \[ U^{(1)}=[\alpha_1u_{1}^{(1)},\cdots, \alpha_r u_{r}^{(1)}],\] 
and $U^{(k)}, 2\leq k\leq t$ are matrices with column vectors given by  $u_1^{(k)},\dots, u_r^{(k)}\in \mathbb R^{n_1}$. By the definition of atomic norm, we have 
\[\|T\|_{\pm}=\sum_{i=1}^r |\alpha_i|.\]
We know that
$T_{I_1, \ldots, I_t} = \bigcirc_{i=1}^t U^{(i)}_{I_i, :}$, where $U^{(i)}_{I_i,:}$ denotes the submatrix of $U^{(i)}$ with the column restricted on $I_i$. 
Therefore $T_{I_1, \ldots, I_t}$ can be written as a linear combination of rank-1 sign tensors with the sum of absolute value of weights given by $\sum_{i=1}^r |\alpha_i|$. 
By the definition of the atomic norm, this is an upper bound, so
$\|T_{I_1, \ldots, I_t}\|_{\pm} \leq \|T\|_{\pm}.$
This proves Claim \eqref{claim:1}.
For Claim \eqref{claim:2},
let \[T = \bigcirc_{i=1}^t T^{(i)}
\quad \text{ and }\quad  S = \bigcirc_{i=1}^t S^{(i)}\]
be the rank $r_1$ and $r_2$ decompositions of $T$ and $S$
that attain their atomic norms. We can write 
\begin{align}
T_{i_1, \ldots, i_t} = 
\sum_{i=1}^{r_1} \alpha_i u^{(1)}_{i_1, i} \cdots u^{(t)}_{i_t, i}, \quad 
    S_{j_1, \ldots, j_t}= 
\sum_{j=1}^{r_2} \beta_j v^{(1)}_{j_1, j} \cdots v^{(t)}_{j_t, j},
\end{align}
 where $v_{j_k,j}^{k}\in \{-1,+1\}$ for $1\leq j_k\leq n_k, 1\leq j\leq r, 1\leq k\leq t$,  \[ V^{(1)}=[\beta_1v_{1}^{(1)},\cdots, \beta_{r_2} v_{r_2}^{(1)}],\] 
and $V^{(k)}, 2\leq k\leq t$ are matrices with column vectors given by  $v_1^{(k)},\dots, v_r^{(k)}$. And by definition, $\|S\|_{\pm}=\sum_{j=1}^{r_2}|\beta_j|$.
Then since
\begin{align*}
(T \otimes S)_{k_1, \ldots, k_t} 
&=
T_{i_1, \ldots, i_t} S_{j_1, \ldots, j_t} =
\left(
\sum_{l=1}^{r_1} U^{(1)}_{i_1, l} \cdots U^{(t)}_{i_t, l} 
\right)
\left(
\sum_{l'=1}^{r_2} V^{(1)}_{j_1, l'} \cdots V^{(t)}_{j_t, l'}
\right) \\
&=
\sum_{l=1}^{r_1}
\sum_{l'=1}^{r_2}
\left( U^{(1)}_{i_1, l} V^{(1)}_{j_1, l'} \right)
\cdots
\left( U^{(t)}_{i_t, l} V^{(t)}_{j_t, l'} \right) \\
&=
\sum_{p=1}^{r_1 r_2}
\left( U^{(1)} \otimes V^{(1)} \right)_{k_1, p}
\cdots
\left( T^{(t)} \otimes V^{(t)} \right)_{k_t, p}
\end{align*}
for $k_s = j_s + m_s (i_s - 1)$
for all $s = 1, \ldots, t$
and $p = l' + r_2 ( l - 1)$,
we have that
$
T \otimes S = \bigcirc_{i=1}^t (U^{(i)} \otimes V^{(i)})$.
This gives a way to write $T \otimes S$ as a weighted sum of rank-$1$ sign tensors with the sum of absolute value of weights given by 
\[ \sum_{i=1}^{r_1}\sum_{j=1}^{r_2} |\alpha_i\beta_j|=\|T\|_{\pm} \|S\|_{\pm}.\]
Therefore 
$\|T \otimes S\|_{\pm} \leq \|T\|_{\pm} \|S\|_{\pm}.$
This completes the proof of  Claim \eqref{claim:2}. 
For Claim \eqref{claim:3}, note that every entry in 
$T *S$ 
appears in 
$T \otimes S$, since
\[
(T *S)_{i_1, \ldots, i_t} 
= 
(T \otimes S)_{i_1 + n_1 (i_1 - 1), \ldots, i_t + n_t(i_t - 1)} .
\]
So we have that $T *S = (T \otimes S)_{I_1, \ldots, I_t}$ for some subsets of indices $I_1,\dots, I_t$,
and by Claim (1),
the result follows. 
Finally, from Claims (2) and (3),
$
    \|T*T\|_{\pm}\leq \|T\otimes T\|_{\pm}\leq \|T\|_{\pm}^2,
$ and Claim \eqref{claim:4} follows.

\subsection{Proof of Lemma \ref{lem:rank-1}}
\label{pf:rank-1}

Since $\|u_i\|_{\infty}\leq 1$, $u_i\in [-1,1]^{n_i}$, where $[-1,1]^{n_i}$ is a convex hull of the set $\{-1,1\}^{n_i}$,  $u_i$ can be written as a convex combination of $\{-1,1\}^{n_i}$ such that 
\[ u_i=\sum_{k\in [2^{n_i}
]}\lambda_k^{(i)} v_k^{i},\]
where $v_k^i, k\in [2^{n_i}]$ are all possible vectors in $\{-1,1\}^{n_i}$ and $\sum_{k} |\lambda_k^{(i)}|=1$. Therefore we have
\[
T=u_1\circ u_2\cdots \circ u_t=\sum_{k_1,\dots,k_t} \lambda_{k_1}^{(1)}\cdots \lambda_{k_t}^{(t)} v_{k_1}^{(1)}\circ \cdots \circ v_{k_t}^{(t)},
\]
which is a decomposition of $T$ as a linear combination of sign rank-1 tensors. So 
\[
\|T\|_{\pm}\leq \sum_{k_1,\dots,k_t} |\lambda_{k_1}^{(1)} \cdots \lambda_{k_t}^{(t)}|\leq 1.
\]

\subsection{Proof of Theorem \ref{thm:atomic_rank}}
\label{pf:atomic_rank}

The lower bound was shown in \cite[Theorem 7]{ghadermarzy2018}. We now focus on the upper bound. When $t=2$, using Grothendieck's inequality, it was shown in \cite[Theorem 7]{heiman2014deterministic} that 
$\|T\|_{\pm} \leq K_G\|T\|_{\max}$. From John's Theorem (see, for example \cite[Corollary 2.2]{rashtchian2016bounded}), $\|T\|_{\max}\leq \sqrt{r} |T|_{\infty}$. This proves \eqref{eq:improve_atomic} when $t=2$.

For $t\geq 3$, we will use induction. Let $T$ be an order $t$ tensor with rank $r$ and $|T|_{\infty}\leq 1$. Then $T$ has the rank-$r$ decomposition as 
\[T=\sum_{j=1}^r v_j^1\circ v_j^2\circ\cdots\circ v_j^t. \]

Matricizing along mode-$1$ we obtain $T_{[1]}\in \mathbb R^{n_1\times (n_2\cdots n_t)}$ such that 
\begin{align*}
    T_{[1]}=\sum_{i=1}^rv_i^1\circ (v_i^2\otimes \cdots\otimes  v_i^t).
\end{align*}
Let $W$ be a $\mathbb R^{(n_2\cdots n_t)\times r}$ matrix such that $W(:,i)=v_i^2\otimes \cdots\otimes  v_i^t$.
By John's Theorem (\cite[Corollary 2.2]{rashtchian2016bounded}), there exists an $S\in \mathbb R^{r\times r}$ such that 
\begin{align}\label{eq:T1x}
  T_{[1]}=X\circ Y,  
\end{align} where $X=V^{(1)}S\in \mathbb R^{n_1\times r}, Y=WS^{-1} \in \mathbb R^{n_2\cdots n_t\times r}$, and $\|X\|_{2,\infty}\leq r^{1/2},\|Y\|_{2,\infty}\leq 1$. This also implies $\|Y\|_{\infty}\leq \|Y\|_{2,\infty}\leq 1$. Since each column of $Y$ is a linear combination of columns of $W$, for some constants $\{\gamma_{ij}\}$,
\[Y(:,i)=\sum_{j=1}^r \gamma_{ij} (v_j^2\otimes \cdots\otimes   v_j^t).\]
Then \[E_i:=\sum_{j=1}^r \gamma_{ij} (v_j^2\circ \cdots\circ   v_j^t)\] is a  order $(t-1)$ tensor of rank at most $r$ with $\|E_i\|_{\infty}\leq 1$. By induction,  we have $\|E_i\|_{\pm} \leq K_G\sqrt{r^{3t-7}}$.
Then by the definition of the atomic norm in \eqref{eq:def_atomic}, there exists a decomposition of $E_i$ such that for some integer $r_i$,
\begin{align*}
E_i=\sum_{j=1}^{r_i}\lambda_j^{(i)}u_{i,j}^2\circ \cdots \circ u_{i,j}^t,
\end{align*}
where \[\sum_{j=1}^{r_i} |\lambda_j^{(i)}|\leq K_G\sqrt{r^{3t-8}},\] and $\|u_{i,j}^d\|_{\infty}\leq 1$ for $2\leq d\leq t$. This gives a decomposition of $T$ such that

\begin{align}\label{eq:decompT}
    T&=\sum_{i=1}^r x_i \circ \left(\sum_{j=1}^{r_i} \lambda_j^{(i)}u_{i,j}^2\circ \cdots \circ u_{i,j}^t\right)
    =\sum_{i=1}^r\sum_{j=1}^{r_i}\lambda_j^{(i)} \|x_i\|_{\infty} \left(\frac{x_i}{\|x_i\|_{\infty}}\right)\circ u_{i,j}^2\circ \cdots \circ u_{i,j}^t,
\end{align}
where $x_i$ is the $i$-th column vector of $X$ and \[\|x_i\|_{\infty} \leq \|x_i\|_2\leq \|X\|_{2,\infty}\leq r^{1/2}.\]
Here we assume all $x_i\not=0$ for $i\in [r]$. Otherwise, we can ignore the terms involving a zero vector.
\eqref{eq:decompT} gives a decomposition of $T$ into a linear combination of $\sum_{i=1}^r r_i$ many rank-1 tensors where each  component vector has $\ell_{\infty}$-norm at most $1$. Therefore by Lemma \ref{lem:rank-1} and  the triangle inequality,
\begin{align*}
\|T\|_{\pm} \leq \sum_{i=1}^r \sum_{j=1}^{r_i} |\lambda_{j}^{(i)}| \|x_i\|_{\infty}\leq r^{3/2} K_G\sqrt{r^{3t-8}}=K_G\sqrt{r^{3t-5}}. 
\end{align*}

% We will use the following lemma from \cite{harris2021deterministic}.
% \begin{lemma}[Lemma SM 1.6 in \cite{harris2021deterministic}]\label{lem:gh23}
% Any order $t$, rank-$r$ tensor $T\in \bigotimes_{i=1}^t \mathbb R^{n_i}$ with $|T|_{\infty}\leq 1$ can be decomposed into $r^{t-1}$ rank-one tensors $T=\sum_{j=1}^{r^{t-1}} u_j^1\circ u_j^2\circ \cdots \circ u_j^t$, where
% \begin{align}\label{eq:a31}
%     \sum_{j=1}^{r^{t-1}} (u_j^k(s))^2\leq r^{t-1} 
% \end{align}
% for any $1\leq k\leq t-1$, $1\leq s\leq n_k$, and 
% \begin{align}\label{eq:a32}
%     \sum_{j=1}^{r^{t-1}} (u_j^t(s))^2\leq r^{t-2}
% \end{align}
% for any $1\leq s\leq n_t$.
% \end{lemma}

\subsection{Proof of Theorem \ref{thm:main3}}
\label{pf:main3}

Following every step in the proof of \cite[Theorem 1.2]{harris2021deterministic}, we have
  \[
  \left| \frac{1}{n^t} \sum_{e \in [n]^t} T_e 
    - \frac{1}{|E|} \sum_{e \in E} T_e \right|
  \leq \frac{2^tn^{t/2}\lambda_2(H)}{|E|}
  \; \| T \|_\pm 
  \]
  holds for any tensor $T$. Now we apply this inequality to
  the tensor of squared residuals
  $R:= (\hat{T} - T ) * (\hat{T} - T).$
  Since we solve for $\hat{T}$
  with equality constraints, we have that
  $R_e = 0$ for all $e \in E$.
  Thus, using Claim \eqref{claim:4} in Theorem \ref{thm:atomic-properties},
  \begin{align}\label{boundlambda1}
  \frac{1}{n^t}\|\hat{T}-T\|_F^2=& 
  \left| \frac{1}{n^{t}} \sum_{e \in [n]^t} R_e \right| \\
  \leq &  \frac{2^tn^{t/2}\lambda_2(H)}{|E|}
  \|R\|_\pm \leq  \frac{2^tn^{t/2}\lambda_2(H)}{|E|} \|\hat{T}-T\|_{\pm}^2\leq \frac{2^{t}n^{t/2}\lambda_2(H)}{|E|} \left(\|\hat T\|_{\pm} +\|T\|_{\pm}\right)^2.\nonumber
  \label{eq:maxinequality}
  \end{align}
  Since $\hat{T}$ is the output of our optimization routine
  and $T$ is feasible, 
  $\| \hat{T} \|_{\pm}\leq \| T \|_{\pm}$.
  This leads to the final result.

\subsection{Proof of Theorem \ref{thm:main_reg}}
\label{pf:main_reg}

Following the same steps in the proof of \cite[Theorem 1.3]{harris2021deterministic},
  \begin{align}
    \left| \frac{1}{n^t} \sum_{e \in [n]^t} T_e 
    - \frac{1}{nd^{t-1}} \sum_{e \in E} T_e \right|
    &\leq 
      2\sum_{j=1}^{2^{t-1}} \frac{(2t-3)\lambda}{4d}=2^{t-2}(2t-3)\frac{\lambda}{d}.
    \end{align}
Define $R=(\hat{T}-T)*(\hat{T}-T)$. Then we have
  \begin{align}\label{boundlambda}
 \frac{1}{n^t}\|\hat{T}-T\|_F^2= \left| \frac{1}{n^{t}} \sum_{e \in [n]^t} R_e \right| 
  \leq 2^{t-2}(2t-3)\frac{\lambda}{d}
  \left(\|\hat{T} \|_{\pm} + \|T \|_{\pm}\right)^2\leq 2^t(2t-3)\frac{\lambda}{d} \|T\|_{\pm}^2.
  \end{align}

\subsection{Proof of Theorem \ref{thm:poisson}}
\label{pf:poisson}

We closely follow the proof of Theorem 3 in \cite{lopez2022zero}. The proof here is modified by using Lemma \ref{thm:atomic_rank} to bound the atomic norm (which improves the rank dependence by a factor of $r$) and applying bounds \eqref{boundlambda1} and \eqref{boundlambda} to replace the uniform random sampling assumption on $E$.

Let $r_{\pm} = \sup_{Z\in S_r(\alpha,\beta)}\|Z\|_{\pm}$. As in the proof of Theorem 3 in \cite{lopez2022zero}, up to equation 3.14, for any $E$ we obtain that with probability greater than $1-2|E|^{-1}$
\[
\frac{1}{|E|} \sum_{e \in E} R_e \leq \frac{128\alpha\sqrt{tn}(2r_{\pm}+2)}{\beta\sqrt{|E|}}\left(\alpha(e^2-2)+3\log_2(|E|)\right)\leq \frac{C\alpha^2t^{3/2}\sqrt{n}r_{\pm}\log_2(n)}{\beta\sqrt{|E|}},
\]
where $R=(\hat{T}-T)*(\hat{T}-T)$ and $C>0$ is an absolute constant. Adding and subtracting $\frac{1}{n^{t}} \sum_{e \in [n]^t} R_e$ to the left-hand side and rearranging, we have shown
\begin{equation}\label{Pbound}
\left|\frac{1}{n^{t}} \sum_{e \in [n]^t} R_e\right| \leq \frac{C\alpha^2t^{3/2}\sqrt{n}r_{\pm}\log_2(n)}{\beta\sqrt{|E|}} + \left|\frac{1}{n^{t}} \sum_{e \in [n]^t} R_e -\frac{1}{|E|} \sum_{e \in E} R_e\right|.
\end{equation}
Bounding the last term as in \eqref{boundlambda1} gives
\[
\left|\frac{1}{n^{t}} \sum_{e \in [n]^t} R_e\right| \leq \frac{C\alpha^2t^{3/2}\sqrt{n}r_{\pm}\log_2(n)}{\beta\sqrt{|E|}} + \frac{2^{t+2}n^{t/2}\lambda_2(H)r_{\pm}^2}{|E|},
\]
which will establish \eqref{eq:Perrorbound2}. To prove \eqref{eq:Perrorbound}, the additional assumptions imposed on $H$ can be used to instead bound the last term of \eqref{Pbound} as in \eqref{boundlambda} and obtain
\[
\left|\frac{1}{n^{t}} \sum_{e \in [n]^t} R_e\right| \leq \frac{C\alpha^2t^{3/2}\sqrt{n}r_{\pm}\log_2(n)}{\beta\sqrt{|E|}} + 2^t(2t-3)\frac{\lambda}{d} r_{\pm}^2.
\]
The proof ends by using $|E|=nd^{t-1}$ and applying Lemma \ref{thm:atomic_rank} to see that $r_{\pm}\leq K_G\alpha\sqrt{r^{3t-5}}$.

\subsection{Square loss experiments}
\label{sec:num_square}

\begin{figure}
\centering
% \includegraphics[width=0.325\linewidth]{eigenvalues_maxqnorm_full_7.png} 
% \hfill
% \includegraphics[width=0.325\linewidth]{eigenvalues_maxqnorm_full_23.png}
% \hfill
% \includegraphics[width=0.325\linewidth]{eigenvalues_maxqnorm_cutoff_23.png}
\includegraphics[width=\linewidth,trim={0 0 0 0 cm},clip]{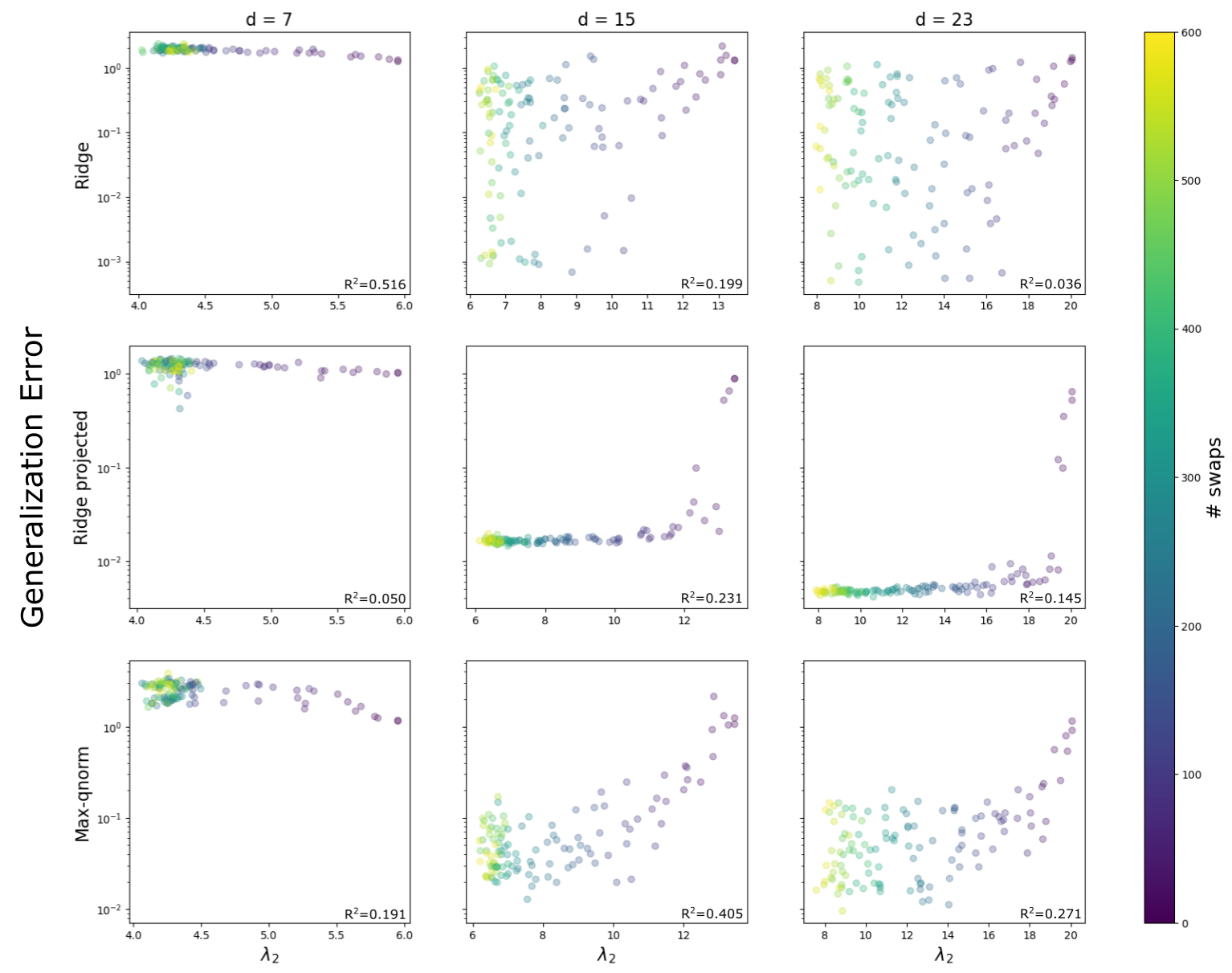}
\caption{
Numerical experiments show that reconstruction error (y-axes, log scale for visualization) correlates with $\lambda_2(G)$ for three different algorithms and varying graph degrees $d=$7, 15, 23.
Coefficients of determination ($R^2$, linear fit of error $\sim \lambda_2$ {\em without} log scaling to be consistent with Fig.~\ref{fig:fig2}) are inlaid, and full regression reports are provided in the supplemental materials.
In the sparsest sampling regime ($d=7$), no method performs well, and the correlation is less consistent.
In the denser regimes, the algorithms perform better, although ridge exhibits very high variance in error, while max-quasinorm and projected ridge are more consistent.
}
\label{fig:full_ranges}
\end{figure}

\begin{figure}
\centering
% \includegraphics[width=0.325\linewidth]{eigenvalues_maxqnorm_full_7.png} 
% \hfill
% \includegraphics[width=0.325\linewidth]{eigenvalues_maxqnorm_full_23.png}
% \hfill
% \includegraphics[width=0.325\linewidth]{eigenvalues_maxqnorm_cutoff_23.png}
\includegraphics[width=\linewidth,trim={0 0 0 0 cm},clip]{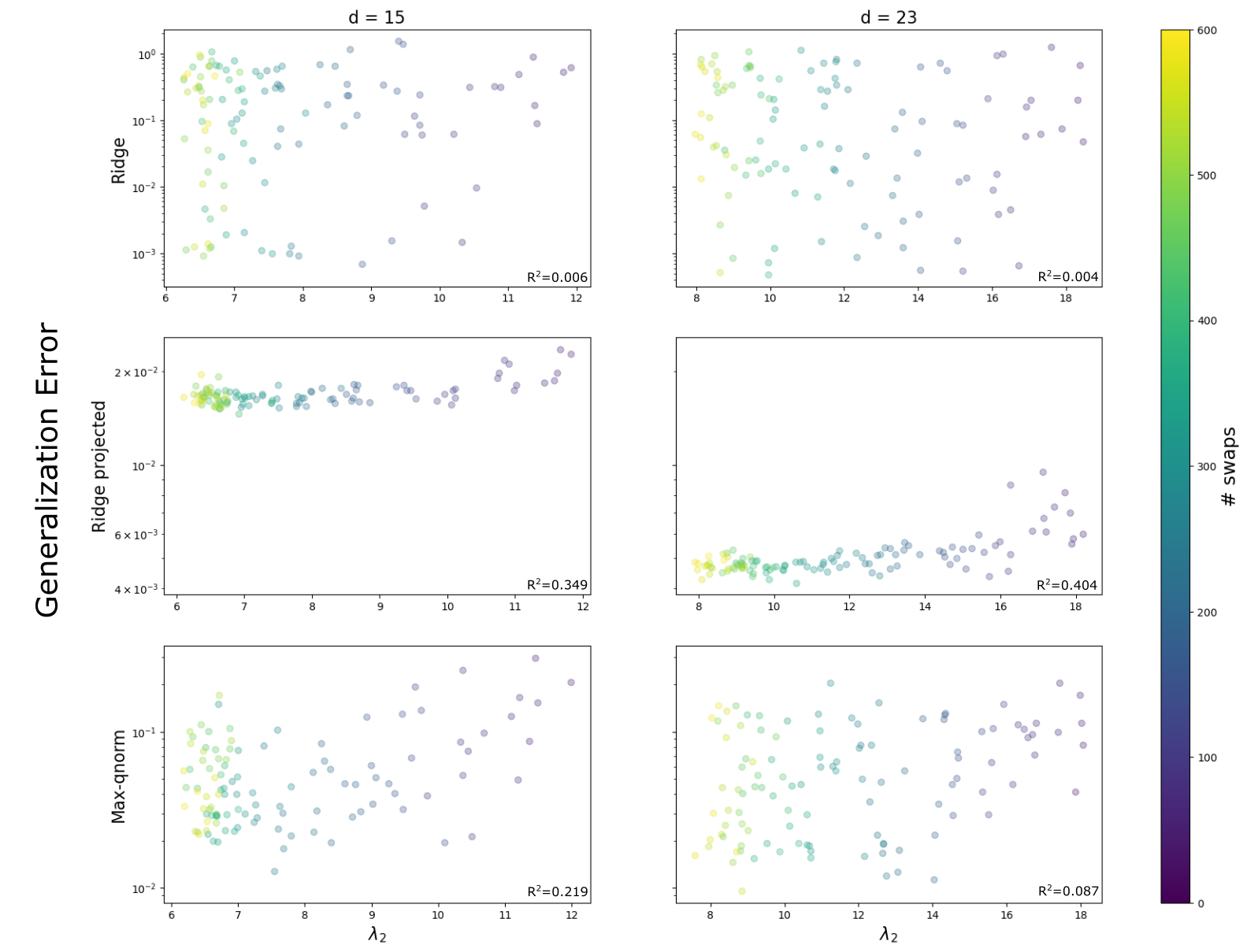}
\caption{
Same as for Fig.~\ref{fig:full_ranges} except cutting off the largest $\lambda_2$ values and just depicting $d=15$, 23.
Numerical experiments show that reconstruction error (y-axes, log scale for visualization) correlates with $\lambda_2(G)$ for three different algorithms and varying graph degrees.
Coefficients of determination ($R^2$, linear fit of error $\sim \lambda_2$ {\em without} log scaling to be consistent with Fig.~\ref{fig:fig2}) are inlaid, and full regression reports are provided in the supplemental materials.
Here, the correlations are generally weaker except for the ridge projected algorithm, which has the best performance.
}
\label{fig:cutoff_ranges}
\end{figure}

Theorems~\ref{thm:main3} and \ref{thm:main_reg} both concern constrained regression where the observations are fit exactly.
However, for noisy data, one can use a square loss and get similar bounds as in \cite{harris2021deterministic}.
We tested the performance of three least-squares tensor completion algorithms for varying spectral gaps.
These differed in their regularization and optimization routines, and we refer to the algorithms as ``ridge,'' ``ridge projected,'' and ``max-quasinorm.''
All of these algorithms are implemented in the code provided at
\url{https://github.com/kamdh/max-qnorm-tensor-completion}

The standard ridge and projected version of it consider a sum of squares penalty on the factor matrices
%\begin{equation*}
$    \sum_{i=1}^t \| U^{(i)} \|_F^2$
%\end{equation*}
as in ridge regression.
The ridge algorithm attempts to solve for
\begin{equation}
    \mbox{Ridge:} \quad
    \min_{T' =U^{(1)} \circ \cdots \circ U^{(t)}}  \|P(T' - T) \|_F^2 
    + \epsilon \sum_{i=1}^t \| U^{(i)} \|_F^2,
\end{equation}
where $T' = U^{(1)} \circ \cdots \circ U^{(t)}$ is the CP decomposition into factor matrices, and $P$ is a projection operator that zeroes out unobserved entries in the data tensor $T$.
The optimization routine performs alternating minimization over the factor matrices--coordinate descent--using the conjugate gradient method. 
This is one of the simpler tensor completion algorithms that one could imagine.

The projected ridge method attempts to solve a constrained version of the ridge problem 
\begin{equation}
       \min_{T' =U^{(1)} \circ \cdots \circ U^{(t)}}  
       \sum_{i=1}^t \| U^{(i)} \|_F^2
       \quad \mbox{s.t.} \quad
       \|P(T' - T) \|_F \leq \delta.
\end{equation}
To deal with the hard constraint on the square residuals, we use an analogous relaxation and variable projection technique as the max-quasinorm algorithm from \cite{harris2021deterministic}.
This leads to the relaxed optimization problem
\begin{align}
    \mbox{Ridge projected:} \quad
    &\min_{T', R}  
    \sum_{i=1}^t \| U^{(i)} \|_F^2
    + \frac{\kappa}{2} \| P( T' - T - R) \|_F^2
    + \beta \| R \|_F^2\\
    &\quad \quad \mbox{s.t.} \quad
    \| R \|_F \leq \delta,    \nonumber
\end{align}
where again $T' =U^{(1)} \circ \cdots \circ U^{(t)}$.
The parameters we used are $\delta = 0.05 \sqrt{|E|}$, $\epsilon = 0.01$, $r_\mathrm{fit} = 10 r$, $\kappa = 100$, $\beta = 1$, {\tt rebalance = True}, {\tt init = 'svdrand'}.
Due to its intriguingly good performance (see below), we plan to study this algorithm in more detail in a future work.

The final algorithm is the max-quasinorm algorithm studied in detail in  \cite{harris2021deterministic}.
This algorithm is the closest to atomic norm minimization that we know of that's also practical.
Atomic norm minimization is challenging since it would require integer optimization.
For completeness, the optimization problem is
\begin{align}
    \mbox{Max-quasinorm:} \quad
    &\min_{T', R}  
    \| T \|_\mathrm{max}
    + \frac{\kappa}{2} \| P( T' - T - R) \|_F^2
    + \beta \| R \|_F^2\\
    &\quad \quad \mbox{s.t.} \quad
    \| P(R) \|_F \leq \delta.    \nonumber
\end{align}
The parameters for the max-quasinorm algorithm were $\delta = 0.05 \sqrt{|E|}$, $\epsilon = 0.01$, $r_\mathrm{fit} = 10 r$, $\kappa = 100$, $\beta = 1$, {\tt rebalance = True}, {\tt init = 'svdrand'}.

The target tensors were size $n=100$, order $t=3$, rank $r=3$, and their factors were generated from a uniform distribution $U[0,1]$ and rescaled to have Hilbert-Schmidt norm $\sqrt{n^t}$.
After fitting $\hat{T}$, we measure generalization error $\| \hat{T} - T \|_F$ using the tensor Frobenius norm (root mean square error).
Due to the normalization of the target tensor, an error of 1 corresponds to 100\% relative error.

We sampled the tensor entries using graph lifting, where a $d$-regular graph $G$ is lifted into a hypergraph by the $t$-path traversal method described in the main text.
We start by taking the $d$-connected ring on $n$ nodes, where each node is connected to its $d$ nearest neighbors with periodic boundary conditions.
This is a deterministic graph with eigenvalue $\lambda_2(G) \approx d-1$ in experiments.
In order to vary $\lambda_2(G)$, we pick edge pairs at random and swap their endpoints.
These edge swaps preserve the degree distribution, but after many swaps, the graph distribution approaches that of the $d$-regular random graph, which has $\lambda_2(G) \approx 2\sqrt{d-1}$, approximately as small as possible.
We take $d \in \{7, 15, 23\}$ to generate hypergraphs with varying proportions 0.5\%, 2.3\%, 5.3\% of observed entries.

Besides the results shown in Figure~\ref{fig:fig2} for $d=15$ and just the max-quasinorm algorithm,
we show supporting results for other parameters and algorithms in Figure~\ref{fig:full_ranges}.
These also show a significant correlation between $\lambda_2(G)$ and the reconstruction error of the tensor for all cases except $d=7$.
In that case, there are too few observations to correctly learn the tensor no matter the gap.
We performed linear regression of error versus $\lambda_2$.
Coefficients of determination are given in the plot, and the full regression reports are given in supplemental tables.
Finally, Figure~\ref{fig:cutoff_ranges} shows the same data over a smaller range of $\lambda_2$.

\subsection{Poisson loss experiments}
\label{sec:num_poi}

We used the Poisson max-likelihood algorithm from \cite{lopez2022zero} with code provided by those authors.
The algorithm was run with its default parameters. The target tensors were size $n=100$, order $t=3$, and rank $r=3$.
Their factors were generated randomly from the uniform distribution and rescaled so the resulting tensor had entries in $[1,6]$.

The sampling was performed without graph lifting.
We found that a regularly spaced ``grid'' tensor had a large $\lambda_2(H)$, and that swapping grid entries with random entries caused the eigenvalue to decrease.
To generate the grid for a particular sampling fraction, we uniformly spaced ones in a length $n^3$ linear array that gets reshaped into an $n \times n \times n$ array.
To vary $\lambda_2(H)$, we then shuffle some fraction of those ones into random locations.
In all experiments, we observe 5\% of the entries and shuffle between 10\% and 100\% of those grid points with random locations.

\begin{figure}
    \centering
    \includegraphics[width=0.48\linewidth]{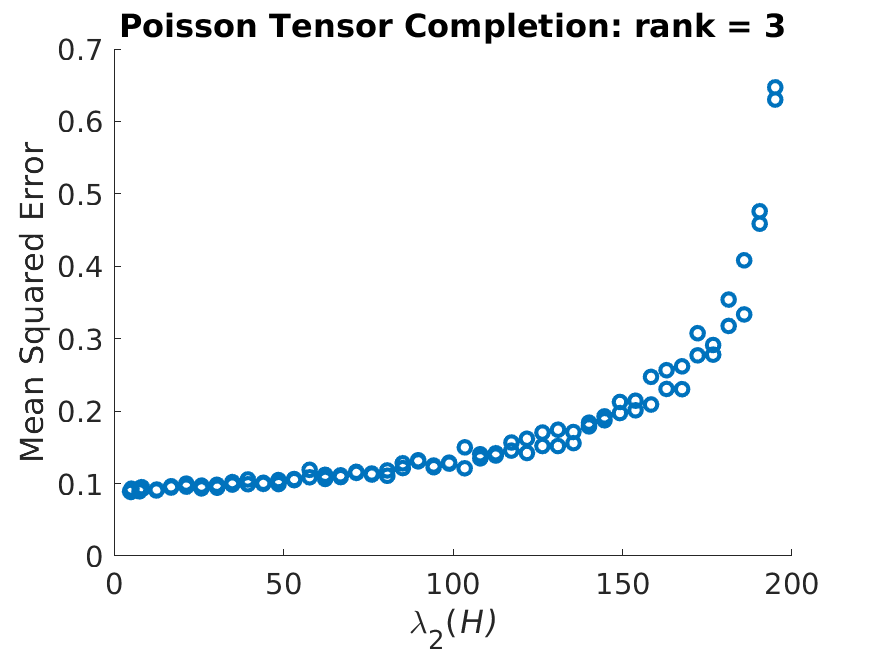}
    \hfill
    \caption{Results of Poisson tensor regression mean squared error for masks with varying $\lambda_2(H)$. A strong correlation is exhibited for target tensor with rank $r=3$.
    }
    \label{fig:poisson}
\end{figure}

To estimate $\lambda_2(H)$, we fit a rank-1 tensor $R$ to $(T_H - \frac{|E|}{n^t}J)$ using alternating least squares, so that
$R \approx (T_H - \frac{|E|}{n^t}J)$.
The Frobenius norm $\|R\|_F$ gives an estimate of the second eigenvalue. Figure~\ref{fig:poisson} shows how the mean squared error $\frac{1}{n^{3}}\|\hat{T} - T\|_F^2$ varies with $\lambda_2(H)$.
Again, we see a strong positive correlation between error and the eigenvalue.

\end{document}